\documentclass[lettersize,journal]{IEEEtran}
\usepackage{amsmath,amsfonts,amssymb}
\usepackage{algorithm}
\usepackage{array}
\usepackage[caption=false,font=normalsize,labelfont=sf,textfont=sf]{subfig}
\usepackage{textcomp}
\usepackage{stfloats}
\usepackage{url}
\usepackage{verbatim}
\usepackage{graphicx}
\usepackage{cite}
\usepackage{graphicx}
\usepackage{amsmath}
\usepackage{amssymb}
\usepackage{booktabs}
\usepackage{amsfonts}
\usepackage{array}
\usepackage{hyperref}
\usepackage{comment}

\usepackage{color}
\usepackage{bbding}
\usepackage{pifont}
\usepackage{wasysym}
\usepackage{cite}
\usepackage{stfloats}

\usepackage{multirow}
\usepackage{makecell}
\usepackage{diagbox}
\usepackage{threeparttable}
\usepackage{algpseudocode}

\usepackage{mathtools}
\usepackage{float}
\hypersetup{hidelinks}

\usepackage{bm}
\hypersetup{
    colorlinks=true,       
    linkcolor=black,        
    filecolor=black,     
    urlcolor=blue,         
    citecolor=black,        
}

\makeatletter
\let\@oldmaketitle\@maketitle
\renewcommand{\@maketitle}{
  \@oldmaketitle
  \centering
  \vspace{1em} 
  \includegraphics[width=\linewidth]{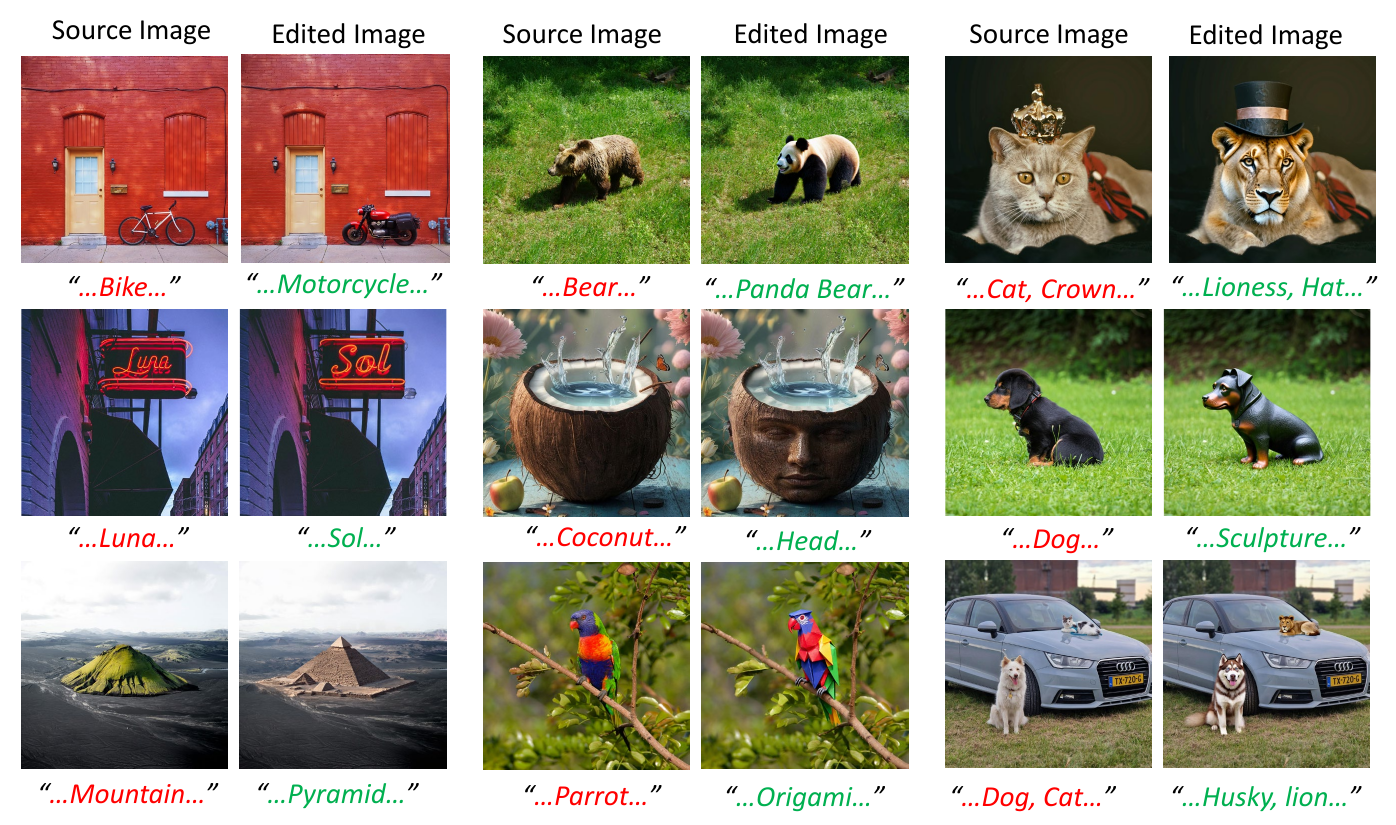}
  \vspace{0.7em} 
  
  \parbox{\linewidth}{
    \small
    Fig. 1.~SAM-Flow achieves excellent semantic editing results while preserving the background of the source image. The red text indicates the source prompt, and the green text indicates the target prompt.
  }
  \vspace{-0.7em} 
  \setcounter{figure}{1}
}
\makeatother

\begin{document}

\title{SAM-Flow: Source-Anchored Masked Flow for Training-Free Image Editing}
\author{
 Haowang Cui,
    Rui Chen\textsuperscript{†},
    Tao Luo,
    Tao Guo,
    Zheng Qin,
  Jiaze Wang%
\thanks{This work was supported by the National Natural Science Foundation of China under Grant 62272341. \textit{(† Corresponding author. Corresponding author: Rui Chen.)}

Haowang Cui, Rui Chen, Tao Guo, Jiaze Wang and Zheng Qin are with Tianjin Key Laboratory of Imaging and Sensing Microelectronic Technology, School of Microelectronics, Tianjin University, Tianjin 300072, China (e-mail: haowangcui@tju.edu.cn; ruichen@tju.edu.cn; guotao\_stu@tju.edu.cn; jiaze\_w@tju.edu.cn; zhengqin58@tju.edu.cn).

Tao Luo is with School of Cyber Security, Tianjin University, Tianjin 300072, China (e-mail: luo\_tao@tju.edu.cn)

}
 }


\maketitle

\begin{abstract}
Training-free image editing has recently attracted increasing attention due to its ability to modify real images using powerful pre-trained diffusion and flow-matching models without additional training. However, existing inversion-based and differential-flow-based methods usually perform global latent transport, which inevitably propagates editing effects to non-target regions and leads to background leakage. To address this problem, we propose SAM-Flow, a source-anchored masked flow framework for localized training-free image editing. Instead of updating the whole latent representation, SAM-Flow first uses a scout image and token-grounded attention maps to localize the editable semantic regions. It then applies differential velocity updates only within these regions, while anchoring the remaining areas to the source-image latent trajectory. To further improve spatial stability and boundary naturalness, we introduce a time-varying source-anchored projection mechanism with dynamic soft masks, transition regions, and temporal mask accumulation. The proposed method is plug-and-play and can be integrated with mainstream flow-matching backbones such as Stable Diffusion 3 and FLUX without any fine-tuning. Extensive qualitative and quantitative experiments demonstrate that SAM-Flow achieves accurate semantic editing while significantly improving background preservation, providing a simple and general localized editing paradigm for training-free image editing. Code is available at: \href{https://github.com/chwbob/Sam-Flow}{\textit{https://github.com/chwbob/Sam-Flow}}.


\end{abstract}

\begin{IEEEkeywords}
Flow matching, training-free image editing, localized editing, background preservation
\end{IEEEkeywords}

\section{Introduction}
\IEEEPARstart{D}{iffusion} models~\cite{DDPM,scorebased_model} and flow-matching methods~\cite{flow_matching,rectified_flow} have significantly advanced text-to-image generation. Recent state-of-the-art generative models, such as Stable Diffusion 3~\cite{SD3} and FLUX~\cite{flux2024}, formulate the generation process as an ordinary differential equation (ODE) that transports a simple noise distribution to a complex natural image distribution, achieving strong visual fidelity and text-prompt alignment. Built upon such powerful generative priors, image editing has become an important downstream task with great practical value \cite{sdedit,imagic,instructpix2pix,prompt_to_prompt_image_editing,p_and_p_image_trans}. Training-free editing methods can perform complex semantic modifications on real images using only text prompts, while avoiding the expensive cost of additional training or fine-tuning \cite{prompt_to_prompt_image_editing,p_and_p_image_trans,null_text_inversion,pnp_inversion,infEdit}.

\begin{figure*}[htb]
    \centering
    \includegraphics[width=\textwidth]{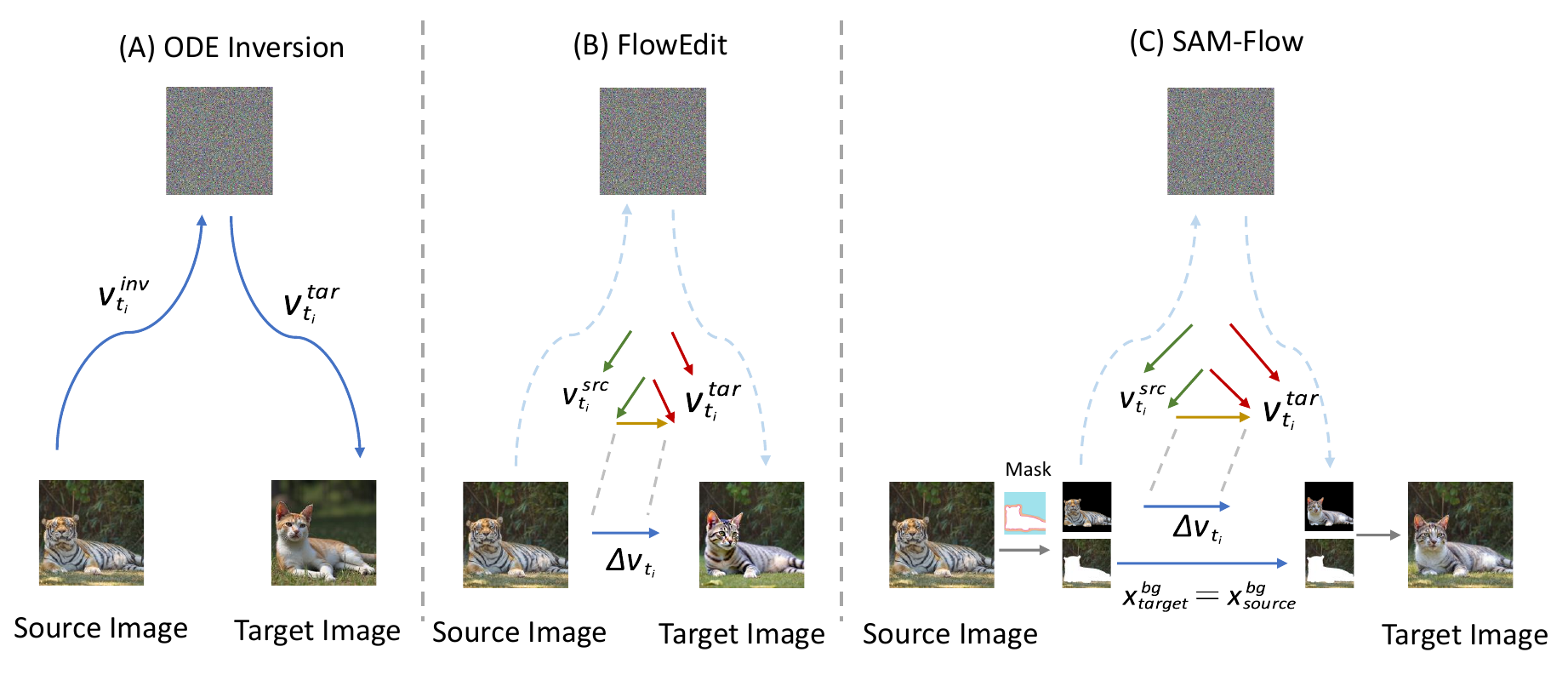}
    \caption{Comparison between SAM-Flow and two mainstream paradigms of training-free image editing.}
    \label{fig4-1}
\end{figure*}

Existing training-free image editing methods can be roughly divided into two paradigms. The first paradigm is based on ODE inversion~\cite{ddpm_noise_inv,null_text_inversion,EDICT,direct_inversion}. These methods invert a source image back to the latent noise distribution through an inversion velocity field $v_{t_i}^{\mathrm{inv}}$, and then perform forward sampling under the guidance of a target velocity field $v_{t_i}^{\mathrm{tar}}$ to achieve editing, as illustrated in Fig.~\ref{fig4-1}(A). The second paradigm is represented by differential-flow-based methods such as FlowEdit~\cite{flowedit}. Instead of conducting costly inversion, FlowEdit directly computes the differential velocity between the target velocity field $v_{t_i}^{\mathrm{tar}}$ and the source velocity field $v_{t_i}^{\mathrm{src}}$:
\begin{equation}
    \Delta v_{t_i} = v_{t_i}^{\mathrm{tar}} - v_{t_i}^{\mathrm{src}},
\end{equation}
and uses it to drive the editing trajectory in the latent space, as shown in Fig.~\ref{fig4-1}(B). Although these two paradigms have different advantages, they share a fundamental limitation: both rely on global distribution transport. Whether the editing trajectory is planned in the inverted noise space or driven by a differential velocity field over the entire latent space, the whole image is updated during the editing process. Such global transport inevitably forces background pixels in the non-edited regions to adapt to the target distribution, resulting in severe and often irreversible edit leakage. This greatly weakens the preservation of source-image characteristics in regions that are supposed to remain unchanged.

To overcome the inherent limitation of global distribution transport, we propose SAM-Flow, a controllable content editing method that shifts the training-free editing paradigm from global distribution mapping to precise localized flow editing, as shown in Fig.~\ref{fig4-1}(C). Instead of blindly applying global velocity updates, SAM-Flow explicitly isolates the semantic target regions and applies differential-flow-based editing only within these localized regions. Meanwhile, the remaining background regions are strictly anchored to the latent features of the source image, aiming to satisfy the constraint:
\begin{equation}
    x_{\mathrm{target}}^{\mathrm{bg}} = x_{\mathrm{source}}^{\mathrm{bg}}.
\end{equation}
In this way, as shown in Fig. 1, SAM-Flow preserves the source image outside the edited region while allowing the target semantics to be effectively generated inside the desired area.

A key challenge for precise localized flow editing lies in the spatial misalignment between the target object in the source image and that in the edited image. To address this problem, we introduce a Token-Grounded Semantic Localization mechanism. Specifically, we first generate a preliminary edited image, termed the scout image, using an unconstrained image editing method. We then extract token-grounded attention maps corresponding to the target semantic tokens from both the source image and the scout image, following the observation that cross-attention maps in text-to-image diffusion models can provide word-level spatial localization cues \cite{prompt_to_prompt_image_editing,tang2023daam,Attend-and-excite,masactrl}. By taking the union of these attention maps, SAM-Flow constructs a spatial mask that covers the complete regions involved in the edit. This scout-and-union strategy effectively reduces artifacts caused by missing mask regions and enables diverse editing operations, including object addition, removal, replacement, and large structural deformation.

Furthermore, to avoid hard blending boundaries and ensure smooth transitions across different ODE time steps, we design a time-varying source-anchored projection mechanism. Unlike simple static masking, SAM-Flow dynamically partitions the union attention map into three regions: the core region, the transition region, and the background region. The boundary threshold, mask softness, and transition radius are adaptively updated along the flow-matching trajectory. At each update step, the differential velocity field is applied only to the core and transition regions, while the latent features of the background region are projected back to the source-image latent trajectory. This mechanism effectively suppresses edit leakage while maintaining the natural coherence of the generated image.

The main contributions of our work are summarized as follows:

\begin{itemize}
    \item \textbf{Localized flow editing framework:} We identify global distribution transport as a fundamental limitation of both inversion-based and differential-flow-based training-free editing. We propose SAM-Flow, which shifts the editing process toward precise localized flow updates and effectively reduces background leakage.

    \item \textbf{Token-grounded semantic localization:} We introduce a scout-and-union attention strategy that combines source-side and scout-side token-grounded responses. This design provides sufficient spatial coverage for object replacement, addition, removal, and large deformation.

    \item \textbf{Source-anchored dynamic projection:} We design a time-varying tri-partite soft mask and integrate it into the ODE solving process. By combining core regions, transition regions, temporal accumulation, and source-latent anchoring, SAM-Flow smoothly blends localized edits while preserving non-edited regions.

    \item \textbf{Plug-and-play generality:} SAM-Flow can be integrated with mainstream flow-matching backbones, such as FLUX~\cite{flux2024} and Stable Diffusion 3~\cite{SD3}, without training or fine-tuning, providing a simple and general localized editing paradigm.
\end{itemize}


\section{Related Work and Preliminaries}

\subsection{Flow Matching Models}

Diffusion models have achieved remarkable progress in text-to-image generation~\cite{DDPM,LDM}. Recently, flow matching and rectified flow have provided an alternative formulation by learning a continuous velocity field that transports samples from a simple prior distribution to the data distribution~\cite{flow_matching,rectified_flow}. In this formulation, the generation process can be described by an ordinary differential equation (ODE):
\begin{equation}
\frac{d x_t}{d t}=v_\theta(x_t,t,p),
\end{equation}
where $x_t$ denotes the latent state at time step $t$, $p$ denotes the text condition, and $v_\theta(\cdot)$ is the velocity field predicted by the pre-trained model. Recent large-scale text-to-image models, such as Stable Diffusion 3 and FLUX, adopt rectified-flow or flow-matching-style formulations and show strong generation quality and text alignment~\cite{SD3,flux2024}. Since the predicted velocity field varies with the text condition, it can also be used to estimate the semantic transport direction between a source prompt and a target prompt, which forms the foundation of flow-based training-free image editing.

\subsection{Inversion-Based Image Editing}

Training-free image editing aims to modify real images using pre-trained generative models without additional model training. A widely used paradigm is inversion-based editing. These methods first invert a source image into a noise code or an intermediate latent trajectory, and then perform forward sampling under the target prompt~\cite{DDIM,null_text_inversion,prompt_to_prompt_image_editing,p_and_p_image_trans,negative_prompt_inversion}. In an ODE-based formulation, given a source image latent $x_{t_0}^{\mathrm{src}}$, inversion estimates a trajectory from the source image state to a noisy latent state. This process can be written as:
\begin{equation}
x_{t_{i+1}}^{\mathrm{inv}}
=
x_{t_i}^{\mathrm{inv}}
+
(t_{i+1}-t_i)v_{t_i}^{\mathrm{inv}},
\end{equation}
where $v_{t_i}^{\mathrm{inv}}$ denotes the inversion velocity field at time step $t_i$. After obtaining the inverted latent state, editing is performed by solving the forward ODE with the target prompt:
\begin{equation}
x_{t_{i-1}}^{\mathrm{tar}}
=
x_{t_i}^{\mathrm{tar}}
+
(t_{i-1}-t_i)v_{t_i}^{\mathrm{tar}},
\end{equation}
where $v_{t_i}^{\mathrm{tar}}=f_\theta(x_{t_i}^{\mathrm{tar}},t_i,p^{\mathrm{tar}})$ is the target velocity field. Inversion-based methods enable real-image editing and can be combined with attention control, feature injection, or prompt optimization~\cite{prompt_to_prompt_image_editing,null_text_inversion,p_and_p_image_trans,pnp_inversion}. However, their performance is sensitive to inversion accuracy. Reconstruction errors introduced during inversion may be amplified during editing. Moreover, since the forward editing trajectory is usually applied to the whole latent representation, non-target regions may also be affected by the target prompt.

\subsection{Inversion-Free Image Editing via Differential Flow}

To avoid explicit inversion, FlowEdit proposes an inversion-free and optimization-free editing framework for pre-trained text-to-image flow models~\cite{flowedit}. Instead of first mapping the source image back to the noise space, FlowEdit directly constructs an editing ODE between the source and target distributions. At each time step $t_i$, it computes the velocity field under the source prompt and the target prompt:
\begin{equation}
v_{t_i}^{\mathrm{src}}
=
f_\theta(x_{t_i}^{\mathrm{src}},t_i,p^{\mathrm{src}}),
\end{equation}
\begin{equation}
v_{t_i}^{\mathrm{tar}}
=
f_\theta(x_{t_i}^{\mathrm{tar}},t_i,p^{\mathrm{tar}}),
\end{equation}
where $p^{\mathrm{src}}$ and $p^{\mathrm{tar}}$ denote the source and target prompts, respectively. The semantic editing direction is then estimated by the differential velocity field:
\begin{equation}
\Delta v_{t_i}
=
v_{t_i}^{\mathrm{tar}}
-
v_{t_i}^{\mathrm{src}}.
\end{equation}
This differential velocity describes the instantaneous transport direction from the source semantics to the target semantics. The editing trajectory is updated as:
\begin{equation}
x_{t_{i-1}}^{\mathrm{edit}}
=
x_{t_i}^{\mathrm{edit}}
+
(t_{i-1}-t_i)\Delta v_{t_i}.
\end{equation}
Compared with inversion-based methods, FlowEdit avoids explicit image inversion and latent optimization, thereby reducing inversion-induced reconstruction errors and improving compatibility with modern flow-matching backbones such as Stable Diffusion 3 and FLUX~\cite{SD3,flux2024,flowedit}. Several recent works have also explored inversion, feature sharing, or trajectory regularization for rectified-flow-based editing~\cite{rfedit,StableFlow,reflex,flowalign}. Among them, FlowEdit is particularly related to our method because it directly uses the velocity difference as the semantic editing force. SAM-Flow inherits this differential-flow formulation, but constrains the differential velocity within localized editable regions rather than applying it to the entire latent representation.

\subsection{Masked Diffusion Editing}

Another line of research introduces spatial constraints into diffusion-based image editing, inpainting, and controllable generation \cite{blended_diffusion,blended_latent_diffusion,glide,repaint,diffedit,Paint_by_Example,inpaint_anything,controlnet,t2i-adapter,gligen}. Masked diffusion editing methods usually require a user-provided or automatically estimated mask to specify the editable region. Blended Diffusion performs local text-driven editing by combining a diffusion prior with CLIP guidance and an ROI mask~\cite{blended_diffusion}. GLIDE demonstrates text-guided image inpainting by regenerating masked regions under textual conditions~\cite{glide}. RePaint uses a pre-trained unconditional diffusion model for image inpainting and repeatedly resamples known regions from the source image during reverse diffusion~\cite{repaint}. DiffEdit further estimates an editing mask by contrasting diffusion predictions under different text prompts and then performs mask-guided semantic editing~\cite{diffedit}. Other inpainting, exemplar-based, segmentation-assisted, or region-aware editing methods also rely on masks or spatial conditions to preserve known regions while synthesizing new content inside the target area \cite{Paint_by_Example,inpaint_anything,sam,controlnet,t2i-adapter,gligen}. These methods show that spatial masks are effective for localizing edits and preserving background regions. However, most of them still depend on external masks, automatically estimated static masks, or hard blending between edited and preserved regions.

\subsection{Limitations of Global Transport and Static Masking}

The above two paradigms reveal two complementary limitations. First, both inversion-based editing \cite{DDIM,null_text_inversion,EDICT,ddpm_noise_inv,pnp_inversion} and inversion-free differential-flow editing \cite{flowedit,infEdit,StableFlow,flowalign} can be viewed as global or weakly localized distribution transport. Although FlowEdit removes the explicit inversion stage, its differential velocity field $\Delta v_{t_i}$ is still applied to the whole latent representation. Therefore, the target prompt may influence not only the desired object but also unrelated background regions, leading to background drift, texture leakage, identity changes, and layout distortion. This problem is especially severe when the target prompt introduces a strong semantic change.

Second, masked diffusion and inpainting methods provide spatial constraints \cite{blended_diffusion,glide,repaint,diffedit,Paint_by_Example,inpaint_anything}, but their masks are often manual, externally obtained, or static. Manual masks increase user burden and are difficult to obtain precisely. Static masks are also insufficient for complex semantic editing because the source and target editable regions may not be spatially aligned. In object replacement, the target object may have a different shape, scale, or position from the source object. In object addition, the target object does not exist in the source image. In object removal, the target prompt may no longer explicitly contain the removed object. Moreover, hard or fixed mask boundaries may produce unnatural transitions between edited and preserved regions.

These observations motivate a localized flow-based editing framework that combines the advantages of inversion-free differential flow and masked editing. Such a framework should preserve the efficient semantic editing direction provided by FlowEdit, while replacing global latent transport with dynamic, semantically localized, and source-anchored masked flow. This is the main motivation of the proposed SAM-Flow framework.


\section{Methodology}

\begin{figure*}[htb]
    \centering
    \includegraphics[width=\linewidth]{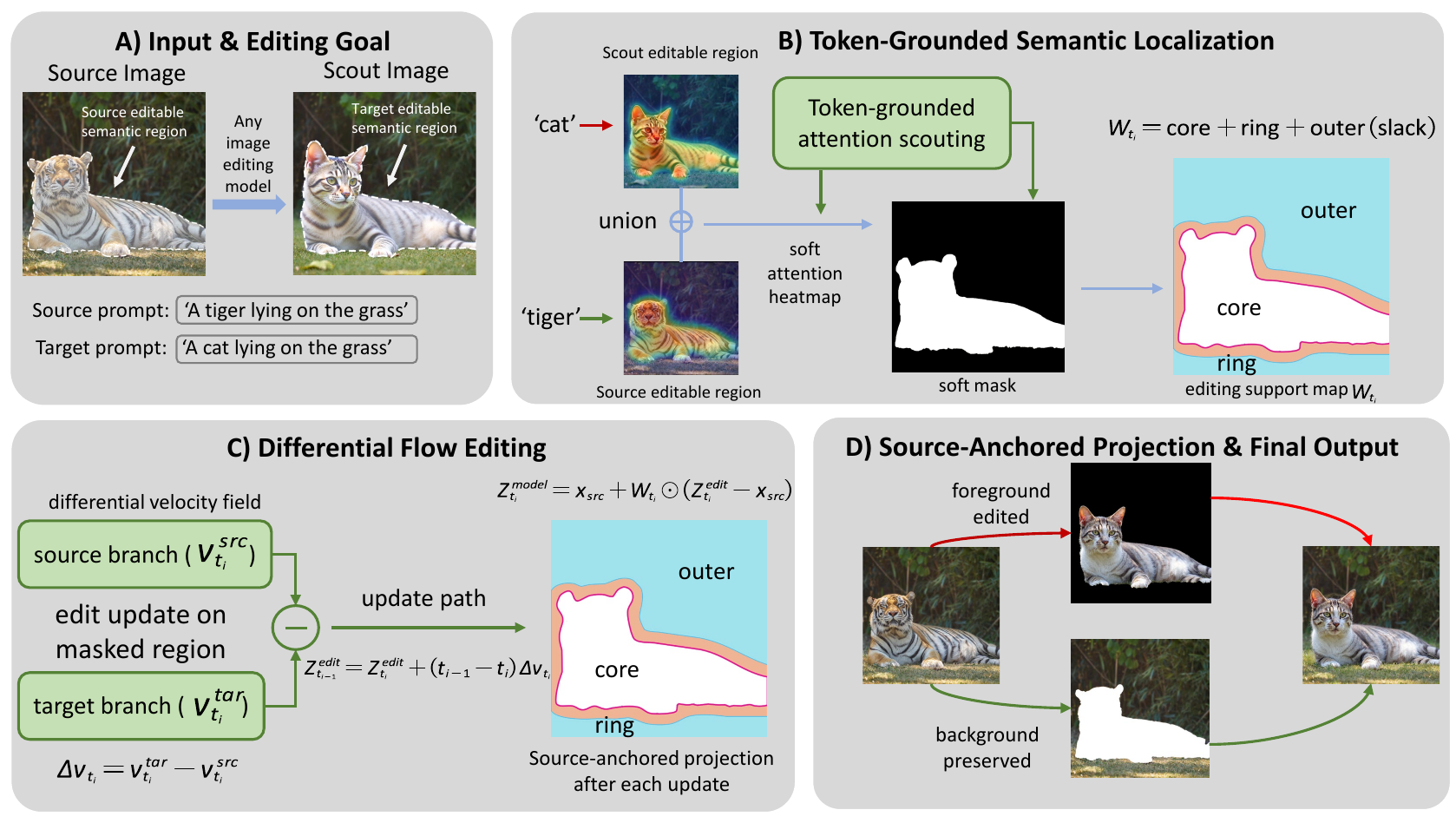}
    \caption{The detailed pipeline of SAM-Flow.}
    \label{fig4.2}
\end{figure*}

SAM-Flow reformulates training-free image editing from a global latent transport problem into a localized semantic flow editing problem. Given a source image $I^{src}$, a source prompt $p^{src}$, and a target prompt $p^{tar}$, where the superscripts $src$ and $tar$ denote the source and target domains, respectively, the goal is to synthesize an edited image that follows $p^{tar}$ while preserving the non-edited content of $I^{src}$. Instead of transporting the whole latent representation from the source distribution to the target distribution, SAM-Flow decomposes the problem into semantic localization, localized differential-flow editing, and source-anchored projection.

As illustrated in Fig.~\ref{fig4.2}, the whole pipeline consists of four stages. First, SAM-Flow generates a preliminary edited image, termed the scout image $I^{scout}$, using an unconstrained editing method. The scout image is not used as the final result, but acts as a semantic probe that reveals where the target concept may appear after editing. Second, token-grounded attention maps are extracted from both the source image and the scout image to construct an editable support region. Third, the semantic editing direction is estimated by the differential velocity between the target and source prompts, following the spirit of FlowEdit~\cite{flowedit}, but this velocity is applied only within the editable support. Finally, the remaining latent regions are repeatedly projected back to the source-image latent trajectory through a time-varying source-anchored projection mechanism. Complete backbone-specific extraction details, the full algorithm, and implementation settings are provided in Appendices~\ref{app:attention}--\ref{app:settings}.

For clarity, we introduce the notation used in this section. The superscript $src$ denotes quantities associated with the source image, $tar$ denotes quantities associated with the target editing branch, $scout$ denotes quantities associated with the scout image, $keep$ denotes quantities associated with the preserved region, $base$ denotes the base attention map, $hard$ and $soft$ denote hard and soft support maps, $model$ denotes the internal state of the ODE solver, $vis$ denotes the visible state used for final decoding, $core$ denotes the core editing region, and $ring$ denotes the transition region. The subscript $t_i$ denotes the $i$-th time step. If an additional subscript $k$ appears, it denotes the $k$-th Monte Carlo sample. Variables with an overline, such as $\bar{W}$, denote temporally accumulated quantities.

\subsection{Token-Grounded Semantic Localization}

The key difficulty of localized training-free image editing is that the editable region is both unknown and dynamically dependent on the target semantics. A mask extracted only from the source image is often incomplete. In object replacement, the new object may differ in shape, scale, and position from the original object. In object addition, the target object does not exist in the source image. In object removal, the target prompt may no longer explicitly contain the removed object. Therefore, SAM-Flow first uses a scout image to estimate the possible target-side semantic region, and then combines it with the source-side semantic region.

Let $E(\cdot)$ denote the VAE encoder~\cite{vae} of the backbone model. The source image and scout image are encoded as:
\begin{equation}\label{eq:encode_src_main}
    z^{src} = E(I^{src}),
\end{equation}
\begin{equation}\label{eq:encode_scout_main}
    z^{scout} = E(I^{scout}).
\end{equation}
The scout branch is used only for spatial localization and does not directly participate in the numerical update of the final editing trajectory. In other words, the scout branch determines where editing may occur, while the subsequent flow editing branch determines how the latent trajectory evolves.

For each editing case, SAM-Flow uses token-level semantic annotations, including a source mask token set, a target mask token set, and optionally an unchanged token set. At the $i$-th time step, we construct noisy latents for the source image and the scout image with the same reference noise:
\begin{equation}\label{eq:noisy_src_main}
    \tilde z^{src}_{t_i} = (1 - t_i)z^{src} + t_i\epsilon^{ref},
\end{equation}
\begin{equation}\label{eq:noisy_scout_main}
    \tilde z^{scout}_{t_i} = (1 - t_i)z^{scout} + t_i\epsilon^{ref},
\end{equation}
where $\epsilon^{ref}\sim\mathcal{N}(0,I)$. Sharing the same noise instance makes the attention responses of the source and scout branches directly comparable at the same time step, which is important for constructing a stable union mask.

SAM-Flow then extracts token-grounded spatial attention maps from the two branches. Let $M^{src}_{t_i}$ denote the normalized attention map obtained from editable source-side tokens in the source branch, and let $M^{tar}_{t_i}$ denote the normalized attention map obtained from editable target-side tokens in the scout branch. The base editing support map is defined as their pixel-wise union:
\begin{equation}\label{eq:base_union_main}
    M^{base}_{t_i} = \max\left(M^{src}_{t_i}, M^{tar}_{t_i}\right).
\end{equation}
This union design covers both the original semantic location and the possible target semantic location. For replacement, $M^{src}_{t_i}$ provides the region of the original object, while $M^{tar}_{t_i}$ provides the possible region of the new object. For addition, the source image may not contain the new object, so the scout branch provides the necessary target-region cue. For removal, the source branch still locates the object to be removed. Thus, replacement, addition, and removal can be handled by the same localization formulation.

SAM-Flow also supports a complementary localization mode for cases where the main subject should remain unchanged while the surrounding environment is edited. Given a preservation map $M^{keep}_{t_i}$ extracted from the unchanged tokens, the editable region is defined as:
\begin{equation}\label{eq:complement_main}
    M^{base}_{t_i}=1-M^{keep}_{t_i}.
\end{equation}
This mode is useful for editing weather, season, background style, or surrounding objects while preserving the main foreground subject.

The localization principle is shared by all backbones, but the attention extraction implementation is model-specific. For FLUX, we intercept scaled dot-product attention in the multimodal DiT~\cite{DIT} and extract the image-to-text response by selecting image queries and text keys from the joint image--text sequence. The responses are averaged over heads and aggregated over middle-to-late layers. Token matching is performed by input-ID subsequence matching with the T5 tokenizer~\cite{t5}, which makes the method robust to words composed of multiple sub-tokens. For Stable Diffusion 3, we use both image-to-text and text-to-image joint attention. Since SD3 uses two CLIP encoders~\cite{clip} and one T5 encoder~\cite{t5}, we build a unified text-weight vector over the concatenated text layout and combine the attention responses from different text encoders. The exact layer ranges, token weights, and aggregation strategy are summarized in Appendix~\ref{app:attention}.

\subsection{Localized Differential Flow Editing}

After the editable support region is obtained, SAM-Flow performs semantic editing using a differential velocity field. Different from FlowEdit, which applies the differential velocity globally, SAM-Flow treats it as a local editing force. The velocity difference determines what semantic change should happen, while the support map and source anchor determine where this change is allowed to propagate.

Let $z^{model}_{t_i}$ denote the editable latent state internally used by the solver at the $i$-th time step, and let $z^{src}$ denote the fixed source-image latent. For the $k$-th Monte Carlo sample, SAM-Flow constructs a source branch and a target branch as:
\begin{equation}\label{eq:src_branch_main}
    z^{src}_{t_i,k} = (1-t_i)z^{src} + t_i\epsilon_k,
\end{equation}
\begin{equation}\label{eq:tar_branch_main}
    z^{tar}_{t_i,k} = z^{model}_{t_i} + z^{src}_{t_i,k} - z^{src},
\end{equation}
where $\epsilon_k\sim\mathcal{N}(0,I)$. The same noise realization is used to construct the two correlated branches. This correlated construction reduces the influence of random noise and makes the difference between model predictions focus more on the semantic displacement induced by the source and target prompts.

The pre-trained flow model predicts the source and target velocity fields as:
\begin{equation}\label{eq:v_src_main}
    v^{src}_{t_i,k}=f_\theta(z^{src}_{t_i,k}, t_i, p^{src}),
\end{equation}
\begin{equation}\label{eq:v_tar_main}
    v^{tar}_{t_i,k}=f_\theta(z^{tar}_{t_i,k}, t_i, p^{tar}).
\end{equation}
The differential editing velocity is then estimated by averaging the velocity differences over $n_{avg}$ samples:
\begin{equation}\label{eq:delta_v_main}
    \Delta v_{t_i}=\frac{1}{n_{avg}}\sum_{k=1}^{n_{avg}}\left(v^{tar}_{t_i,k}-v^{src}_{t_i,k}\right).
\end{equation}
This formulation inherits the inversion-free advantage of FlowEdit~\cite{flowedit}. It does not require mapping the source image back to the Gaussian noise space and does not rely on optimization. However, in SAM-Flow, $\Delta v_{t_i}$ is not regarded as a global transport direction. It is first used to obtain an unconstrained editing candidate:
\begin{equation}\label{eq:edit_update_main}
    z^{edit}_{t_{i-1}}=z^{edit}_{t_i}+(t_{i-1}-t_i)\Delta v_{t_i}.
\end{equation}
If this candidate were passed directly to the next step, the method would degenerate into global differential-flow editing and would inherit the background leakage problem. Therefore, SAM-Flow uses $z^{edit}_{t_{i-1}}$ only as a candidate state and subsequently projects it through the dynamic source-anchored mask.

This separation is essential: the differential velocity field provides semantic editability, while the projection mechanism provides locality and preservation. In particular, source and target prompts are allowed to induce strong semantic changes inside the support region, but their influence is suppressed in non-edited regions by repeatedly anchoring these regions to the source latent. This design allows SAM-Flow to preserve the plug-and-play and inversion-free nature of FlowEdit while avoiding its global latent transport behavior.

\subsection{Time-Varying Source-Anchored Projection}

A static binary mask is insufficient for localized flow editing. At early flow-matching steps, attention responses are usually broad and uncertain, because the latent state still contains high-level structural ambiguity. At later steps, attention maps become sharper and more object-specific. A hard mask may therefore miss target regions at early stages or create unnatural boundaries at later stages. SAM-Flow addresses this issue with a time-varying source-anchored projection mechanism that combines dynamic soft masks, transition regions, and temporal accumulation.

Starting from the base attention map $M^{base}_{t_i}$, we first apply Gaussian smoothing and min-max normalization. We then define a dynamic quantile threshold:
\begin{equation}\label{eq:q_schedule_main}
    q_i=q_{start}+(q_{end}-q_{start})\rho_i,
\end{equation}
\begin{equation}\label{eq:beta_schedule_main}
    \beta_i=\mathrm{clip}\!\left(\mathrm{Quantile}(M^{base}_{t_i},q_i),0.1,0.6\right),
\end{equation}
where $\rho_i=i/T$ denotes the normalized time-step progress, $T$ is the number of time steps, and $q_{start}$ and $q_{end}$ control the evolution of the threshold. The threshold is relatively permissive at early steps and becomes more selective later.

We further introduce a time-varying sharpening coefficient:
\begin{equation}\label{eq:alpha_schedule_main}
    \alpha_i=\alpha_{min}+(\alpha_{max}-\alpha_{min})\rho_i,
\end{equation}
and construct the soft semantic response:
\begin{equation}\label{eq:soft_mask_main}
    M_{t_i}=\sigma\!\left(\alpha_i(M^{base}_{t_i}-\beta_i)\right),
\end{equation}
where $\sigma(\cdot)$ denotes the logistic sigmoid function. As $\alpha_i$ increases, the mask boundary becomes progressively sharper, which matches the coarse-to-fine evolution of the flow trajectory.

The core editing region is defined as:
\begin{equation}\label{eq:core_region_main}
    C_{t_i}=\mathbf{1}[M_{t_i}>\tau_{core}],
\end{equation}
where $\tau_{core}$ is the core threshold. To avoid abrupt boundary discontinuities, we dilate the core region using a time-varying radius $r_i$. Let $D_{t_i}$ denote the dilated result. The transition region is defined as:
\begin{equation}\label{eq:ring_region_main}
    R_{t_i}=\mathrm{clamp}(D_{t_i}-C_{t_i},0,1).
\end{equation}
The transition region provides a soft blending band between the strongly edited core and the preserved background. We then construct the hard support map and the final soft support map as:
\begin{equation}\label{eq:hard_support_main}
    W^{hard}_{t_i}=\mathrm{clamp}\!\left(M_{t_i}+w_{ring}R_{t_i}(1-M_{t_i}),0,1\right),
\end{equation}
\begin{equation}\label{eq:soft_support_main}
    W^{soft}_{t_i}=s+(1-s)W^{hard}_{t_i},
\end{equation}
where $w_{ring}$ controls the transition-region enhancement and $s$ is a small relaxation factor. The relaxation term allows minimal flexibility in non-edited regions, which avoids excessively rigid projection artifacts while still preserving the source content.

Attention maps may fluctuate across time steps, especially for thin structures or newly generated objects. To stabilize the editable support, SAM-Flow maintains accumulated support states:
\begin{equation}\label{eq:accum_soft_main}
    \bar W^{soft}_{t_i}=\max\left(\bar W^{soft}_{t_{i+1}},W^{soft}_{t_i}\right),
\end{equation}
\begin{equation}\label{eq:accum_hard_main}
    \bar W^{hard}_{t_i}=\max\left(\bar W^{hard}_{t_{i+1}},W^{hard}_{t_i}\right),
\end{equation}
\begin{equation}\label{eq:accum_core_main}
    \bar C_{t_i}=\max\left(\bar C_{t_{i+1}},C_{t_i}\right),
\end{equation}
\begin{equation}\label{eq:accum_ring_main}
    \bar R_{t_i}=\max\left(\bar R_{t_{i+1}},R_{t_i}\right).
\end{equation}
Since the accumulation is implemented by element-wise maximum, a region once identified as editable will not be suddenly removed in subsequent steps. This monotonic support memory is particularly important for object addition and large deformation, where the target semantic region may emerge gradually along the trajectory.

After the unconstrained candidate $z^{edit}_{t_{i-1}}$ is obtained, SAM-Flow fuses it with a latent anchor to construct the new model state. Let $a^{model}_{t_i}$ denote the anchor for the solver state. In the default setting, $a^{model}_{t_i}$ is set to the source-image latent reference. The source-anchored model-state projection is:
\begin{equation}\label{eq:model_projection_main}
    z^{model}_{t_{i-1}}=a^{model}_{t_i}+\rho\,\bar W^{soft}_{t_i}\odot\left(z^{edit}_{t_{i-1}}-a^{model}_{t_i}\right),
\end{equation}
where $\rho$ controls the allowed deviation from the anchor and $\odot$ denotes element-wise multiplication. Inside the support region, $\bar W^{soft}_{t_i}$ is large and the latent state can move toward the edited candidate. Outside the support region, $\bar W^{soft}_{t_i}$ is small and the state is pulled back toward the source anchor. Therefore, background preservation is not imposed only at the final output, but repeatedly enforced during the ODE solving process.

In addition to the internal model state, SAM-Flow maintains a visible latent state $z^{vis}_{t_i}$ for final image decoding. We first define three spatial weights based on the accumulated core and transition regions:
\begin{equation}\label{eq:core_mask_final_main}
    \mathrm{core}_{t_i}=\bar C_{t_i},
\end{equation}
\begin{equation}\label{eq:ring_mask_final_main}
    \mathrm{ring}_{t_i}=\mathrm{clamp}\left(\bar R_{t_i}(1-\bar C_{t_i}),0,1\right),
\end{equation}
\begin{equation}\label{eq:outer_mask_final_main}
    \mathrm{outer}_{t_i}=\mathrm{clamp}\left(1-\mathrm{core}_{t_i}-\mathrm{ring}_{t_i},0,1\right).
\end{equation}
The strong core state and the conservative transition state are constructed as:
\begin{equation}\label{eq:core_vis_main}
    z^{core}_{t_{i-1}}=a^{model}_{t_i}+\rho\,\bar W^{soft}_{t_i}\odot\left(z^{edit}_{t_{i-1}}-a^{model}_{t_i}\right),
\end{equation}
\begin{equation}\label{eq:ring_vis_main}
    z^{ring}_{t_{i-1}}=a^{vis}_{t_i}+\rho\,(\gamma\bar W^{hard}_{t_i})\odot\left(z^{edit}_{t_{i-1}}-a^{vis}_{t_i}\right),
\end{equation}
where $a^{vis}_{t_i}$ is the visible-state anchor and $\gamma$ weakens the editing strength in the transition region. Finally, the visible latent state is obtained by:
\begin{equation}\label{eq:visible_composition_main}
    z^{vis}_{t_{i-1}}=\mathrm{core}_{t_i}\odot z^{core}_{t_{i-1}}
    +\mathrm{ring}_{t_i}\odot z^{ring}_{t_{i-1}}
    +\mathrm{outer}_{t_i}\odot a^{vis}_{t_i}.
\end{equation}
The core region performs the main semantic editing, the transition region alleviates boundary artifacts, and the outer region preserves the source anchor. After the full latent trajectory has evolved, the final image is decoded from $z^{vis}$ rather than from the unconstrained state $z^{edit}$. This ensures that the final output reflects the complete source-anchored composition instead of a global differential-flow editing trajectory. The algorithmic summary is provided in Appendix~\ref{app:algorithm}.

\begin{figure*}[htb]
    \centering
    \includegraphics[width=\linewidth]{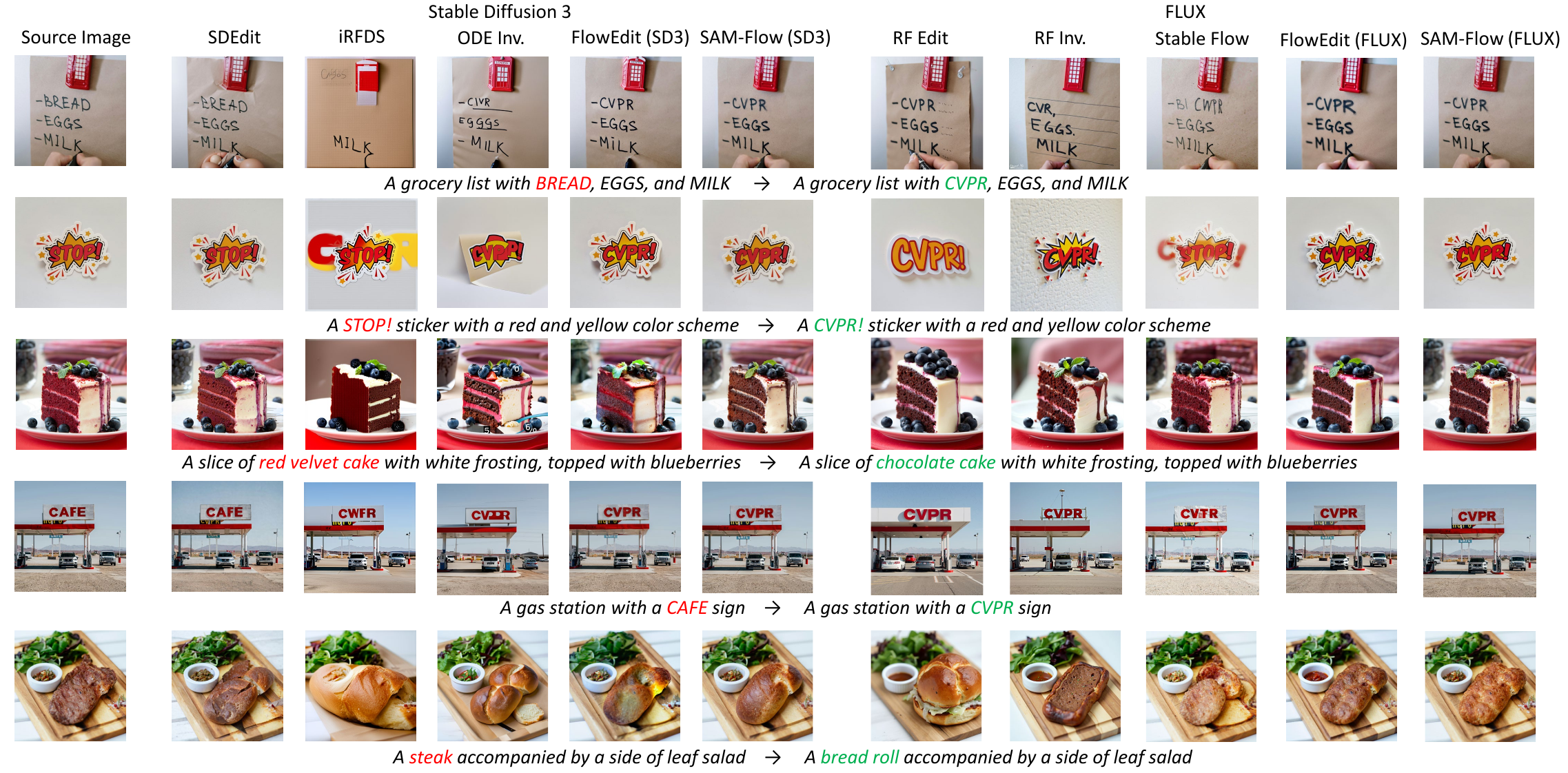}
    \caption{Qualitative comparison between SAM-Flow and other methods. The red text indicates the source prompt, and the green text indicates the target prompt.}
    \label{fig4.5}
\end{figure*}

\section{Experiments}
\label{sec:experiments}

\subsection{Experimental Setup}

We evaluate SAM-Flow on the DIV2K-based~\cite{DIV2K} editing benchmark used in the FlowEdit setup and extend its prompt annotations to support token-grounded localization. Each record contains a source image with a resolution of $1024\times1024$, a source prompt, multiple target prompts, and token-level annotations, including \texttt{source\_mask\_token}, \texttt{target\_mask\_token}, and \texttt{unchanged\_token}. The benchmark covers object replacement, material/style transfer, text replacement, attribute modification, multi-object editing, object addition/removal, and background transformation.

We conduct experiments using Stable Diffusion 3 Medium~\cite{SD3} and FLUX.1-dev~\cite{flux2024} as backbone models. For each test sample, we first generate a scout image using FlowEdit, then execute the SAM-Flow editing pipeline. We compare with SDEdit~\cite{sdedit}, iRFDS~\cite{irfds}, ODE Inversion, FlowEdit~\cite{flowedit}, RF Inversion~\cite{rf_inversion}, RF Edit~\cite{rfedit}, and StableFlow~\cite{StableFlow}. We report CLIP-T~\cite{clip} for instruction faithfulness and DINO score~\cite{dino} for overall visual consistency. To evaluate background preservation, we use the final SAM-Flow mask to segment foreground and background regions, and compute PSNR, SSIM~\cite{SSIM}, and LPIPS~\cite{lpips} only on the background region. Complete hyperparameters, baseline settings, and evaluation details are provided in Appendix~\ref{app:settings}.

\subsection{Main Results}

\begin{figure}[htb]
    \centering
    \includegraphics[width=0.8\linewidth]{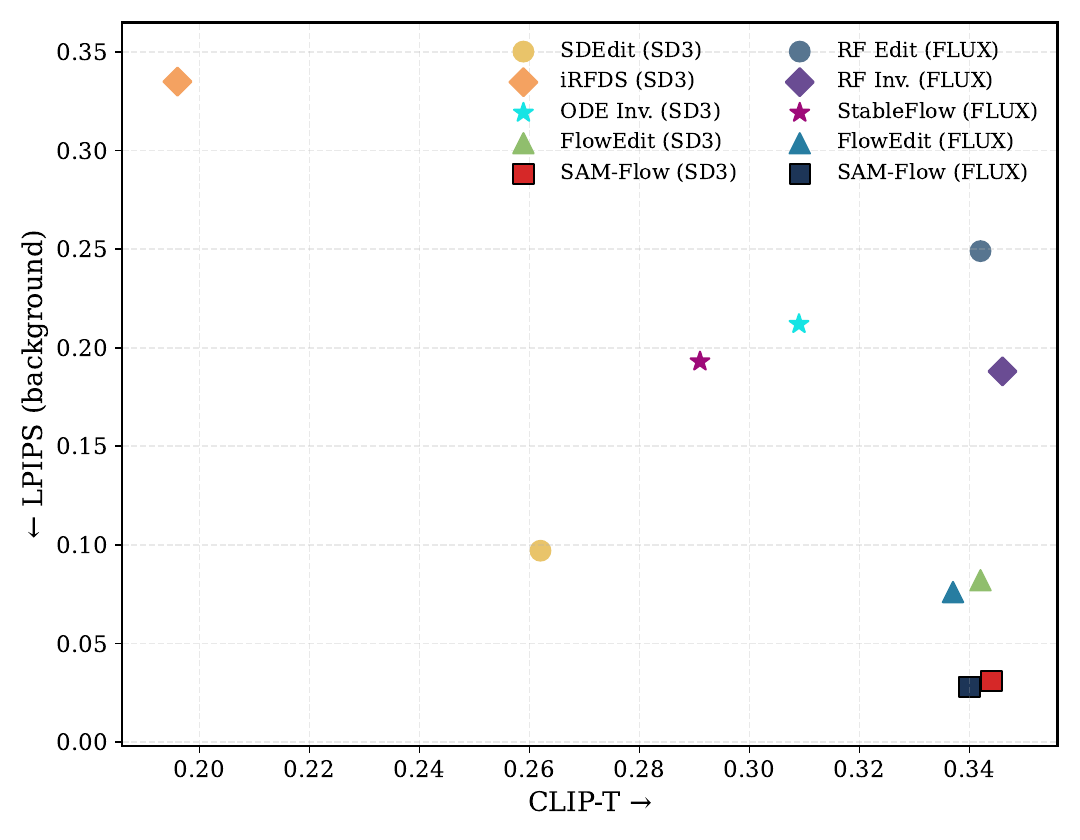}
    \caption{Quantitative comparison between SAM-Flow and other methods.}
    \label{fig4.6}
\end{figure}

Fig.~\ref{fig4.5} shows qualitative comparisons between SAM-Flow and baseline methods. SAM-Flow generates accurate target semantics while significantly reducing background drift. Compared with SDEdit and iRFDS, it produces clearer and more controllable edits. Compared with RF Edit, RF Inversion, and FlowEdit, it better preserves the source layout and non-edited regions.

The quantitative results show that the advantage of SAM-Flow does not lie in maximizing a single metric alone. Instead, it achieves a better overall trade-off between instruction following and background fidelity. Fig.~\ref{fig4.6} visualizes this trade-off using CLIP-T as the horizontal axis and background LPIPS as the vertical axis. Both SAM-Flow (SD3) and SAM-Flow (FLUX) are located in the lower-right region, corresponding to higher CLIP-T and lower background LPIPS. The full quantitative table, mask visualization, and user study results are provided in Appendix~\ref{app:main_results}. The user study further confirms that SAM-Flow receives the highest overall preference among the compared methods.

\subsection{Ablation Study}

\begin{figure}[htb]
    \centering
    \includegraphics[width=\linewidth]{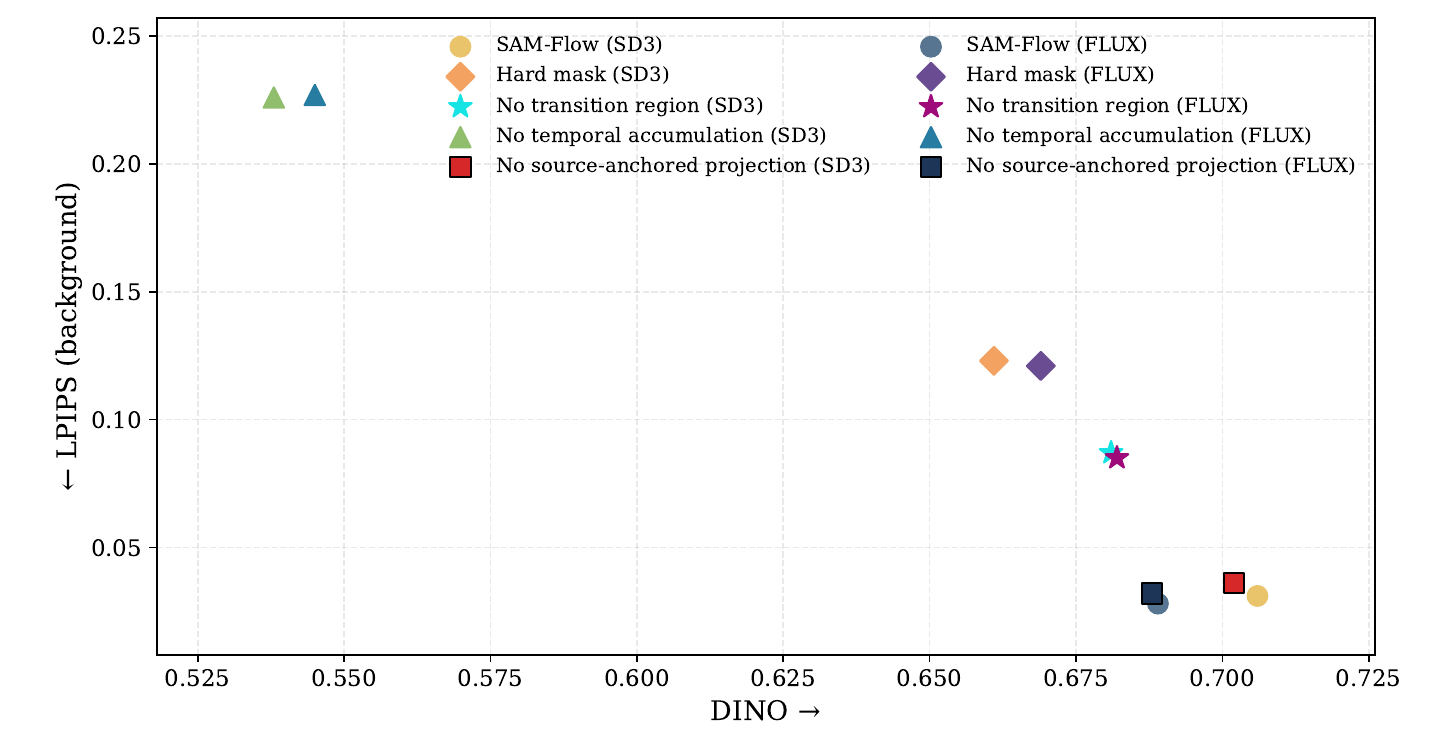}
    \caption{Ablation study visualized by the trade-off between DINO and background LPIPS. A better variant should move toward the lower-right region, indicating better global consistency and lower background perceptual error.}
    \label{fig4.8}
\end{figure}

To analyze the contribution of each component, we design four ablation variants: \textbf{hard mask}, \textbf{no transition region}, \textbf{no temporal accumulation}, and \textbf{no source-anchored projection}. These variants respectively test the effects of soft support, boundary transition, cross-time-step mask memory, and true source-latent anchoring. Detailed settings, equations, qualitative examples, and complete quantitative tables are provided in Appendix~\ref{app:ablation}.

As shown in Fig.~\ref{fig4.8}, the full SAM-Flow obtains the most favorable trade-off on both SD3 and FLUX. Hard masks and removing transition regions weaken boundary naturalness and increase background perceptual error. Removing temporal accumulation causes the most significant degradation because the editable support becomes unstable across time steps. Removing source-anchored projection also increases background drift, confirming that true source-latent anchoring is essential for preserving non-edited regions.


\section{Conclusion}
\label{sec:conclusion}

In this paper, we proposed SAM-Flow, a training-free image editing method that reformulates image editing as localized semantic flow editing rather than global distribution transport. By using a scout image and token-grounded attention maps, SAM-Flow identifies editable semantic regions and applies differential flow updates only within these regions. A time-varying source-anchored projection mechanism with soft support, transition regions, and temporal accumulation further improves boundary smoothness and background preservation. Experiments on Stable Diffusion 3 and FLUX demonstrate that SAM-Flow achieves a favorable balance between semantic consistency, visual quality, and source-image fidelity, providing a simple and general localized editing paradigm for training-free image editing.


\bibliographystyle{IEEEtran}
\bibliography{bibtex/bib/refMS}

\clearpage
\onecolumn
\pagenumbering{arabic} 
\setcounter{page}{1}
\renewcommand{\thepage}{A-\arabic{page}} 

\appendices

\setcounter{figure}{0} 
\renewcommand{\thefigure}{A\arabic{figure}}

\setcounter{table}{0}
\renewcommand{\thetable}{A\arabic{table}}

\setcounter{equation}{0}
\renewcommand{\theequation}{A\arabic{equation}}

\setcounter{algorithm}{0}
\renewcommand{\thealgorithm}{A\arabic{algorithm}}

\section{Backbone-Specific Token-Grounded Attention Extraction}
\label{app:attention}

The main paper describes the general scout-and-union localization principle. This appendix provides the implementation details for different backbone models. The scout image is only used for spatial localization and does not directly participate in the numerical update of the final editing trajectory. In other words, the scout branch is responsible for locating where editing may occur, while the subsequent flow editing process determines how the latent trajectory evolves.

For each editing sample, SAM-Flow receives token-level semantic annotations, including a source mask token set and a target mask token set. At the $i$-th time step, noisy latents for the source image and the scout image are constructed using the same reference noise:
\begin{equation}\label{eq:noisy_src_app}
    \tilde z^{src}_{t_i} = (1 - t_i) z^{src} + t_i \epsilon^{ref},
\end{equation}
\begin{equation}\label{eq:noisy_scout_app}
    \tilde z^{scout}_{t_i} = (1 - t_i) z^{scout} + t_i \epsilon^{ref}.
\end{equation}
Sharing the same noise instance ensures that the spatial responses of the source branch and the scout branch are directly comparable at the same time step.

For FLUX, we intercept the scaled dot-product attention in the multimodal DiT~\cite{DIT} network and extract the image-to-text attention response. Specifically, after determining the joint image--text sequence, we select image queries and text keys to compute the corresponding attention matrix. The responses are averaged over different attention heads and further aggregated over several middle-to-late layers. Token matching is implemented by subsequence matching over the input IDs of the T5 tokenizer~\cite{t5}, which makes the method robust to words composed of multiple sub-tokens.

For Stable Diffusion 3, we use both image-to-text and text-to-image joint attention. Let $A_{IT}$ denote the attention response from image tokens to text tokens, and let $A_{TI}$ denote the attention response from text tokens to image tokens. We construct token-weighted spatial responses from both directions and fuse them into a unified localization map. Since Stable Diffusion 3 uses two CLIP text encoders~\cite{clip} and one T5~\cite{t5} text encoder, we first build a text-weight vector over the concatenated unified text layout, and then compute the final spatial response map by combining $A_{IT}$ and $A_{TI}$. Together, these operations constitute the token-grounded semantic localization stage of SAM-Flow.

The union-based localization mode is defined as:
\begin{equation}\label{eq:base_union_app}
    M^{base}_{t_i} = \max \left(M^{src}_{t_i}, M^{tar}_{t_i}\right),
\end{equation}
where $M^{src}_{t_i}$ denotes the normalized attention map obtained from editable source-side tokens, and $M^{tar}_{t_i}$ denotes the normalized attention map obtained from editable target-side tokens in the scout branch. In addition, SAM-Flow supports a complementary localization mode for cases where the main subject should remain unchanged while the surrounding environment is edited:
\begin{equation}\label{eq:complement_base_map_app}
    M^{base}_{t_i} = 1 - M^{keep}_{t_i},
\end{equation}
where $M^{keep}_{t_i}$ denotes the attention map obtained from preservation tokens.

\section{Full Differential-Flow Formulation}
\label{app:differential_flow}

This appendix provides the complete differential-flow formulation used by SAM-Flow. The main paper emphasizes the localized nature of SAM-Flow, while here we give the explicit latent construction used at each ODE step.

Let $z^{src}$ be the VAE latent of the source image and let $z^{model}_{t_i}$ be the model state at time step $t_i$. For the $k$-th averaging sample, we draw $\epsilon_k\sim\mathcal{N}(0,I)$ and construct:
\begin{equation}\label{eq:app_src_branch_full}
    z^{src}_{t_i,k}=(1-t_i)z^{src}+t_i\epsilon_k,
\end{equation}
\begin{equation}\label{eq:app_tar_branch_full}
    z^{tar}_{t_i,k}=z^{model}_{t_i}+z^{src}_{t_i,k}-z^{src}.
\end{equation}
The source branch follows the noisy marginal of the source image. The target branch is constructed by translating the current editable state with the same noise offset used in the source branch. This correlated branch construction follows the intuition that subtracting two velocity predictions under matched noise perturbations suppresses random noise components and highlights the semantic direction induced by the prompt change.

The source and target velocities are predicted by the frozen flow model:
\begin{equation}\label{eq:app_v_src_full}
    v^{src}_{t_i,k}=f_\theta(z^{src}_{t_i,k},t_i,p^{src}),
\end{equation}
\begin{equation}\label{eq:app_v_tar_full}
    v^{tar}_{t_i,k}=f_\theta(z^{tar}_{t_i,k},t_i,p^{tar}).
\end{equation}
The differential velocity is:
\begin{equation}\label{eq:app_delta_v_full}
    \Delta v_{t_i}=\frac{1}{n_{avg}}\sum_{k=1}^{n_{avg}}\left(v^{tar}_{t_i,k}-v^{src}_{t_i,k}\right).
\end{equation}
The unconstrained candidate update is:
\begin{equation}\label{eq:app_candidate_update_full}
    z^{edit}_{t_{i-1}}=z^{edit}_{t_i}+(t_{i-1}-t_i)\Delta v_{t_i}.
\end{equation}
In FlowEdit, this candidate state is directly propagated to the next step. In SAM-Flow, it is treated only as a semantic candidate and is subsequently filtered through the dynamic source-anchored projection. This distinction is the main reason SAM-Flow can preserve background regions more reliably than global differential-flow editing.

From an implementation perspective, $n_{avg}$ controls the number of velocity evaluations averaged at each time step. A larger value may improve stability but increases inference cost. In our experiments, we follow the practical setting used by the FlowEdit-style backbone configurations and use the hyperparameters reported in Appendix~\ref{app:settings}. The localized projection is independent of the specific value of $n_{avg}$ and can therefore be plugged into different flow-matching backbones.

\section{Dynamic Mask Construction and Source-Anchored Projection}
\label{app:mask_projection}

A fixed mask is insufficient for training-free image editing, because attention distributions are usually broad and uncertain at early time steps, while gradually becoming more concentrated at later time steps. SAM-Flow explicitly designs the projection mechanism as a time-dependent process. As generation proceeds, the blur strength is gradually reduced, the quantile threshold is gradually increased, the sigmoid sharpening strength is progressively enhanced, and the transition radius is gradually shrunk.

Starting from the base attention map $M^{base}_{t_i}$, SAM-Flow first applies Gaussian smoothing and min-max normalization. We then define a dynamic quantile threshold:
\begin{equation}\label{eq:q_schedule}
    q_i = q_{start} + (q_{end} - q_{start}) \rho_i,
\end{equation}
\begin{equation}\label{eq:beta_schedule}
    \beta_i = \mathrm{clip}\!\left(\mathrm{Quantile}(M^{base}_{t_i}, q_i),\, 0.1,\, 0.6\right),
\end{equation}
where $q_i$ denotes the quantile ratio used at the $i$-th time step, $q_{start}$ and $q_{end}$ denote the starting and ending values of the quantile schedule, respectively, $\rho_i=i/T$ denotes the normalized time-step progress, and $T$ is the total number of time steps. The final threshold at the $i$-th time step is denoted by $\beta_i$.

We further define a time-varying sharpening coefficient:
\begin{equation}\label{eq:alpha_schedule}
    \alpha_i = \alpha_{min} + (\alpha_{max} - \alpha_{min}) \rho_i,
\end{equation}
and construct the soft mask:
\begin{equation}\label{eq:soft_mask}
    M_{t_i} = \sigma\!\left(\alpha_i \left(M^{base}_{t_i} - \beta_i\right)\right),
\end{equation}
where $\sigma(\cdot)$ denotes the logistic sigmoid function. As $\alpha_i$ increases, the mask boundary becomes progressively sharper.

The core region is defined as:
\begin{equation}\label{eq:core_region}
    C_{t_i} = \mathbf{1}[M_{t_i} > \tau_{core}],
\end{equation}
where $\tau_{core}$ is the core threshold. To reduce discontinuities caused by abrupt boundary changes, we further dilate the core region with a time-varying radius $r_i$. Let $D_{t_i}$ denote the dilated result. The transition region is defined as:
\begin{equation}\label{eq:ring_region}
    R_{t_i} = \mathrm{clamp}(D_{t_i} - C_{t_i},\, 0,\, 1).
\end{equation}
This transition region provides a smoother blending band between the strongly edited region and the preserved region.

The raw support map is then obtained by:
\begin{equation}\label{eq:raw_support}
    W^{hard}_{t_i} = \mathrm{clamp}\!\left(M_{t_i} + w_{ring} R_{t_i}(1 - M_{t_i}),\, 0,\, 1\right),
\end{equation}
and the final soft support map is defined as:
\begin{equation}\label{eq:soft_support}
    W^{soft}_{t_i} = s + (1-s)W^{hard}_{t_i},
\end{equation}
where $w_{ring}$ denotes the transition-region enhancement weight, and $s\in[0,1]$ is a global relaxation factor that leaves a small amount of flexibility for non-edited regions. In this way, SAM-Flow constructs a tri-partite support structure consisting of the core region, the transition region, and the outer preservation region.

To improve temporal stability, SAM-Flow maintains accumulated support states over time:
\begin{equation}\label{eq:accum_soft}
    \bar W^{soft}_{t_i} = \max\left(\bar W^{soft}_{t_{i+1}},\, W^{soft}_{t_i}\right),
\end{equation}
\begin{equation}\label{eq:accum_hard}
    \bar W^{hard}_{t_i} = \max\left(\bar W^{hard}_{t_{i+1}},\, W^{hard}_{t_i}\right),
\end{equation}
\begin{equation}\label{eq:accum_core}
    \bar C_{t_i} = \max\left(\bar C_{t_{i+1}},\, C_{t_i}\right),
\end{equation}
\begin{equation}\label{eq:accum_ring}
    \bar R_{t_i} = \max\left(\bar R_{t_{i+1}},\, R_{t_i}\right).
\end{equation}
Since the accumulation is implemented by element-wise maximum, once a region is identified as editable at a certain time step, it will not be suddenly removed in subsequent time steps.

After the unconstrained editing state $z^{edit}_{t_{i-1}}$ is updated, SAM-Flow fuses it with a latent anchor to construct the new solver state:
\begin{equation}\label{eq:model_projection_app}
    z^{model}_{t_{i-1}} = a^{model}_{t_i} + \rho \,\bar W^{soft}_{t_i} \odot \left(z^{edit}_{t_{i-1}} - a^{model}_{t_i}\right).
\end{equation}
In the default setting, $a^{model}_{t_i}$ is set to the source-image latent reference. Inside the support region, the latent variable is allowed to move along $z^{edit}_{t_{i-1}}$. Outside the support region, the latent variable is pulled back toward the source-image anchor.

For final decoding, SAM-Flow constructs a visible latent state. We first define three spatial weights:
\begin{equation}\label{eq:core_mask_final}
    \mathrm{core}_{t_i} = \bar C_{t_i},
\end{equation}
\begin{equation}\label{eq:ring_mask_final}
    \mathrm{ring}_{t_i} = \mathrm{clamp}\left(\bar R_{t_i}(1-\bar C_{t_i}),\,0,\,1\right),
\end{equation}
\begin{equation}\label{eq:outer_mask_final}
    \mathrm{outer}_{t_i} = \mathrm{clamp}\left(1-\mathrm{core}_{t_i}-\mathrm{ring}_{t_i},\,0,\,1\right).
\end{equation}
The strong editing state and conservative transition state are:
\begin{equation}\label{eq:core_vis}
    z^{core}_{t_{i-1}} = a^{model}_{t_i} + \rho \,\bar W^{soft}_{t_i} \odot \left(z^{edit}_{t_{i-1}} - a^{model}_{t_i}\right),
\end{equation}
\begin{equation}\label{eq:ring_vis}
    z^{ring}_{t_{i-1}} = a^{vis}_{t_i} + \rho \, (\gamma \bar W^{hard}_{t_i}) \odot \left(z^{edit}_{t_{i-1}} - a^{vis}_{t_i}\right),
\end{equation}
where $\gamma$ denotes the transition-region blending coefficient. Finally, the visible latent state is obtained by:
\begin{equation}\label{eq:visible_composition_app}
    z^{vis}_{t_{i-1}} = \mathrm{core}_{t_i} \odot z^{core}_{t_{i-1}}
    + \mathrm{ring}_{t_i} \odot z^{ring}_{t_{i-1}}
    + \mathrm{outer}_{t_i} \odot a^{vis}_{t_i}.
\end{equation}

\section{Full Algorithm}
\label{app:algorithm}

\begin{algorithm}[htb]
\small
\caption{SAM-Flow for training-free image editing}
\label{alg:samflow_compact}
\begin{algorithmic}
    \renewcommand{\algorithmicrequire}{\textbf{Input:}}
    \renewcommand{\algorithmicensure}{\textbf{Output:}}
    \Require Source image $X^{src}$, source prompt $p^{src}$, target prompt $p^{tar}$, source token set $\mathcal{T}^{src}$, target token set $\mathcal{T}^{tar}$, optional preservation token set $\mathcal{T}^{keep}$, time steps $\{t_i\}_{i=0}^{T}$, number of averaging samples $n_{avg}$
    \Ensure Edited image $X^{tar}$

    \State $X^{scout} \gets \mathrm{ScoutEdit}(X^{src}, p^{src}, p^{tar})$
    \State $Z^{src} \gets E(X^{src}), \quad Z^{scout} \gets E(X^{scout})$
    \State $Z^{edit}_{t_T}, Z^{model}_{t_T}, Z^{vis}_{t_T} \gets Z^{src}$
    \State $\bar W^{soft}, \bar W^{hard}, \bar C, \bar R \gets 0$

    \For{$i = T, T-1, \dots, 1$}
        \State Sample $\epsilon^{ref} \sim \mathcal{N}(0, I)$
        \State $\hat{Z}^{src}_{t_i} \gets (1-t_i)Z^{src} + t_i \epsilon^{ref}$
        \State $\hat{Z}^{scout}_{t_i} \gets (1-t_i)Z^{scout} + t_i \epsilon^{ref}$
        
        \If{$\mathcal{T}^{keep} \neq \varnothing$}
            \State $M^{base}_{t_i} \gets 1 - \mathrm{TokenAttention}(\hat{Z}^{src}_{t_i}, p^{src}, \mathcal{T}^{keep})$
        \Else
            \State $M^{base}_{t_i} \gets \max \left( \begin{aligned}
                &\mathrm{TokenAttention}(\hat{Z}^{src}_{t_i}, p^{src}, \mathcal{T}^{src}), \\
                &\mathrm{TokenAttention}(\hat{Z}^{scout}_{t_i}, p^{tar}, \mathcal{T}^{tar})
            \end{aligned} \right)$
        \EndIf

        \State $(\bar W^{soft}_{t_i}, \bar W^{hard}_{t_i}, \bar C_{t_i}, \bar R_{t_i}) \gets \mathrm{DynamicPartition}(M^{base}_{t_i}, t_i)$

        \State $\Delta V_{t_i} \gets \frac{1}{n_{avg}} \sum_{k=1}^{n_{avg}} \left( V^{tar}_{t_i,k} - V^{src}_{t_i,k} \right)$
        \Statex \hspace{1.5em} where $\epsilon_k \sim \mathcal{N}(0, I)$, and
        \Statex \hspace{1.5em} $Z^{src}_{t_i,k} = (1-t_i)Z^{src} + t_i \epsilon_k,\; Z^{tar}_{t_i,k} = Z^{model}_{t_i} + Z^{src}_{t_i,k} - Z^{src}$
        \Statex \hspace{1.5em} $V^{\{src,tar\}}_{t_i,k} = f_{\theta}(Z^{\{src,tar\}}_{t_i,k}, p^{\{src,tar\}})$

        \State $Z^{edit}_{t_{i-1}} \gets Z^{edit}_{t_i} + (t_{i-1} - t_i)\Delta V_{t_i}$
        \State $Z^{model}_{t_{i-1}} \gets \mathrm{SourceAnchor}(Z^{edit}_{t_{i-1}}, Z^{src}, \bar W^{soft}_{t_i})$
        \State $Z^{vis}_{t_{i-1}} \gets \mathrm{VisibleCompose}(Z^{edit}_{t_{i-1}}, Z^{src}, \dots)$
    \EndFor

    \State \Return $D(Z^{vis}_{t_0})$
\end{algorithmic}
\end{algorithm}

Full algorithm is shown in Algorithm~\ref{alg:samflow_compact}. Overall, the algorithm can be described as: localize first, edit second, and anchor throughout the process. The scout image addresses spatial misalignment, the differential flow provides the semantic editing direction, and the time-varying source-anchored projection keeps the edit localized and visually consistent.

\section{Implementation Details and Hyperparameters}
\label{app:settings}

\subsection{Benchmark}

We evaluate SAM-Flow on the DIV2K-based~\cite{DIV2K} editing benchmark used in the FlowEdit setup, and further extend its prompt annotations to support token-grounded localization. Each record in our prompt file contains a source image with a resolution of $1024\times1024$, a source prompt, multiple target prompts, and a target code for each editing case. On top of the original prompt annotations, we additionally label three token fields for every target case: \texttt{source\_mask\_token}, \texttt{target\_mask\_token}, and \texttt{unchanged\_token}. These annotations support source-target union localization for replacement/addition/removal edits, as well as complement localization for keep-subject/background-only edits. The data loader normalizes these per-target token lists and expands them to match all target prompts in each record. The benchmark covers object replacement, material/style transfer, text replacement, attribute modification, multi-object editing, and background transformation.

\subsection{Hyperparameters}

\begin{table}[h]
\centering
\caption{Main hyperparameter settings of SAM-Flow.}
\label{tab:samflow_settings}
{
\begin{tabular}{lcc}
\toprule
Setting & SAM-Flow (SD3) & SAM-Flow (FLUX) \\
\midrule
Backbone & Stable Diffusion 3 Medium & FLUX.1-dev \\
Steps $T$ & 50 & 28 \\
$n_{avg}$ & 8 & 8 \\
Source guidance & 1.5 & 1.5 \\
Target guidance & 13.5 & 10.5 \\
$n_{max}$ & 33 & 26 \\
Reference noise seed & 999 & 999 \\
Blur $\sigma$ & 2.0 $\rightarrow$ 0.6 & 2.0 $\rightarrow$ 0.6 \\
Quantile $q$ & 0.6 $\rightarrow$ 0.8 & 0.6 $\rightarrow$ 0.8 \\
Sigmoid sharpening strength $\alpha$ & 5.0 $\rightarrow$ 20.0 & 5.0 $\rightarrow$ 20.0 \\
$\tau_{core}$ & 0.5 & 0.85 \\
Ring radius & 6 $\rightarrow$ 3 & 6 $\rightarrow$ 3 \\
$w_{ring}$ & 0.5 & 0.35 \\
Slack & 0.03 & 0.05 \\
Anchor strength $\rho$ & 1.0 & 1.0 \\
Ring blend $\gamma$ & 0.5 & 0.35 \\
Attention reduction & sym & image-to-text \\
Normalization & full & --- \\
Selected layers & 8--17 & 4--18 \\
Token match mode & ids & input-id subsequence \\
Extra text weights & $w_{clip}=1.0$, $w_{t5}=0.3$ & --- \\
\bottomrule
\end{tabular}}
\end{table}

\begin{table}[h]
\centering
\caption{Main hyperparameter settings for baseline methods.}
\label{tab:baseline_settings}
{
\begin{tabular}{ll}
\toprule
Method & Settings \\
\midrule
SDEdit & steps $=50$, $n_{max}=20$, strength $=0.4$, CFG $=13.5$ \\
iRFDS & official default settings \\
ODE Inv. & steps $=50$, $n_{max}=33$, CFG$_{tar}=13.5$, CFG$_{src}=3.5$ \\
FlowEdit (SD3) & steps $=50$, $n_{max}=33$, CFG$_{tar}=13.5$, CFG$_{src}=3.5$, $n_{avg}=8$ \\
RF Inversion & steps $=28$, starting time $=0$, stopping time $=0.28$, $\eta=0.9$ \\
RF Edit & steps $=30$, guidance $=2$, injection $=2$ \\
StableFlow & official default settings \\
FlowEdit (FLUX) & steps $=28$, $n_{max}=24$, CFG$_{tar}=5.5$, CFG$_{src}=1.5$, $n_{avg}=8$ \\
\bottomrule
\end{tabular}}
\end{table}

Detailed hyperparameters used for SAM-Flow and baseline methods in this paper are shown in Table~\ref{tab:samflow_settings} and Table~\ref{tab:baseline_settings}. All experiments are conducted on a single NVIDIA RTX PRO 6000 GPU. For fair comparison, we evaluate against SDEdit~\cite{sdedit}, iRFDS~\cite{irfds}, ODE Inversion, FlowEdit~\cite{flowedit}, RF Inversion~\cite{rf_inversion}, RF Edit~\cite{rfedit}, StableFlow~\cite{StableFlow}, and FlowEdit (FLUX)~\cite{flowedit}. We report CLIP-T~\cite{clip}, full-image DINO score~\cite{dino}, and background-region PSNR, SSIM~\cite{SSIM}, and LPIPS~\cite{lpips}. The final saved SAM-Flow mask at the last editing step is used to segment foreground and background regions.

\section{Additional Main Results and User Study}
\label{app:main_results}

\subsection{Mask Visualization}

\begin{figure*}[h]
    \centering
    \includegraphics[width=0.9\linewidth]{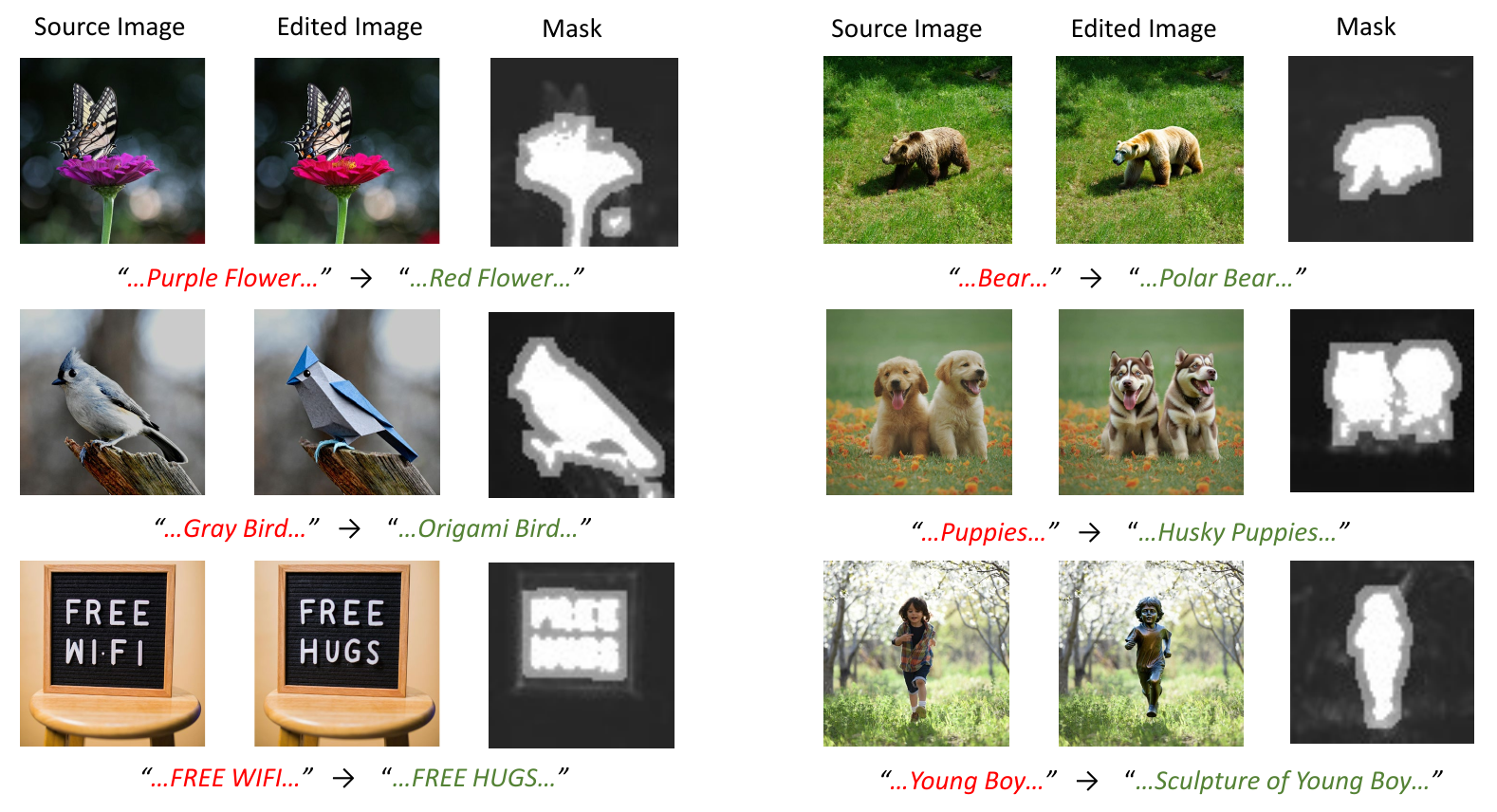}
    \caption{Attention mask results of SAM-Flow. The red text indicates the source prompt, and the green text indicates the target prompt.}
    \label{fig4.4}
\end{figure*}

Fig.~\ref{fig4.4} shows the final masks used by SAM-Flow. The attention masks effectively cover the foreground regions that truly require editing, while leaving non-edited regions available for source anchoring.

\subsection{Full Quantitative Results}

\begin{table}[h]
\centering
\caption{Quantitative comparison between SAM-Flow and other methods.}
\label{tab:main_results}
{
\begin{tabular}{lccccc}
\toprule
Method & CLIP-T $\uparrow$ & DINO $\uparrow$ & PSNR $\uparrow$ & SSIM $\uparrow$ & LPIPS $\downarrow$ \\
\midrule
SDEdit~\cite{sdedit} & 0.262 & \textbf{0.724} & 26.57 & 0.876 & 0.097 \\
iRFDS~\cite{irfds} & 0.196 & 0.494 & 12.09 & 0.433 & 0.335 \\
ODE Inv. & 0.309 & 0.541 & 13.89 & 0.505 & 0.212 \\
FlowEdit (SD3)~\cite{flowedit} & 0.342 & 0.696 & 24.08 & 0.844 & 0.082 \\
SAM-Flow (SD3) & 0.344 & 0.706 & \textbf{29.31} & \textbf{0.925} & 0.031 \\
RF Edit~\cite{rfedit} & 0.342 & 0.614 & 14.82 & 0.693 & 0.249 \\
RF Inv.~\cite{rf_inversion} & \textbf{0.346} & 0.572 & 16.55 & 0.661 & 0.188 \\
StableFlow~\cite{StableFlow} & 0.291 & 0.520 & 18.67 & 0.676 & 0.193 \\
FlowEdit (FLUX)~\cite{flowedit} & 0.337 & 0.686 & 23.94 & 0.863 & 0.076 \\
SAM-Flow (FLUX) & 0.340 & 0.689 & 28.02 & 0.902 & \textbf{0.028} \\
\bottomrule
\end{tabular}}
\end{table}

Full quantitative results are shown in Table~\ref{tab:main_results}. Overall, SAM-Flow achieves a favorable trade-off by maintaining strong instruction-following ability and overall visual quality while significantly improving background fidelity.

\subsection{User Study}

\begin{table}[h]
\centering
\caption{User study results of SAM-Flow on a 10-point scale.}
\label{tab:user_study}
{
\begin{tabular}{lccc}
\toprule
Method & Semantic faithfulness $\uparrow$ & Background preservation $\uparrow$ & Overall score $\uparrow$ \\
\midrule
SDEdit & 4.2 & 8.7 & 6.4 \\
iRFDS & 3.1 & 3.2 & 2.4 \\
ODE Inv. & 5.4 & 3.5 & 3.3 \\
FlowEdit (SD3) & 8.1 & 7.5 & 7.6 \\
SAM-Flow (SD3) & \textbf{8.4} & 9.0 & 9.1 \\
RF Edit & 7.6 & 5.2 & 5.4 \\
RF Inv. & 7.3 & 4.4 & 5.7 \\
StableFlow & 5.1 & 6.0 & 5.8 \\
FlowEdit (FLUX) & 8.2 & 7.9 & 8.3 \\
SAM-Flow (FLUX) & 8.2 & \textbf{8.8} & \textbf{9.2} \\
\bottomrule
\end{tabular}}
\end{table}

We conduct a small-scale user study involving 20 participants, the results are shown in Table~\ref{tab:user_study}. The participants are asked to rate the editing results on a 10-point scale from three aspects: semantic faithfulness, background preservation, and overall quality. The results show that among methods based on Stable Diffusion 3, SAM-Flow (SD3) achieves the highest overall score. Across all methods, SAM-Flow (FLUX) obtains the best overall score.

\section{Ablation Study Details}
\label{app:ablation}

To analyze the contribution of each component in SAM-Flow, we design four ablation variants: \textbf{hard mask}, \textbf{no transition region}, \textbf{no temporal accumulation}, and \textbf{no source-anchored projection}. These variants do not simply remove arbitrary parts of the framework. Instead, each variant modifies a specific key component in the mask-guided editing process while keeping the remaining pipeline unchanged.

For clarity, let $t_i$ denote the $i$-th time step. Let $M_{t_i}$ denote the step-wise soft semantic response map obtained after dynamic thresholding and sigmoid sharpening. Let $C_{t_i}$ denote the core-region mask obtained by binarizing $M_{t_i}$ with the core threshold, and let $R_{t_i}$ denote the transition-region mask constructed around the core region. Furthermore, let $W_{t_i}^{\mathrm{hard}}$ denote the hard support map, and let $W_{t_i}^{\mathrm{soft}}$ denote the soft support map. In terms of latent representation, let $z_{t_i}^{\mathrm{edit}}$ denote the editing latent that is not constrained by anchoring, let $z_{t_i}^{\mathrm{model}}$ denote the model latent state passed to the next denoising step, let $z_{t_i}^{\mathrm{vis}}$ denote the visible latent state used for final image decoding, and let $z^{\mathrm{src}}$ denote the latent representation of the source image encoded by the VAE.

The full SAM-Flow method adopts a time-varying soft partition strategy, together with temporal accumulation and source-latent anchoring. Its model-state update is formulated as:
\begin{equation}\label{eq:ab_full_model}
    z_{t_{i-1}}^{\mathrm{model}} = z^{\mathrm{src}} + \rho \,\bar W_{t_i}^{\mathrm{soft}} \odot \left(z_{t_{i-1}}^{\mathrm{edit}} - z^{\mathrm{src}}\right),
\end{equation}
where $\odot$ denotes element-wise multiplication. Inside the support region, the latent variable is allowed to move toward the edited result. Outside the support region, the latent variable is pulled back toward the source-image latent.

\paragraph{Hard-mask variant.}
The hard-mask variant disables the soft support map mechanism. It sets the transition region to zero:
\begin{equation}\label{eq:ab_hard_mask_r}
    R_{t_i} = 0,
\end{equation}
and directly uses the core-region mask as both the hard and soft support maps:
\begin{equation}\label{eq:ab_hard_mask_whard}
    W_{t_i}^{\mathrm{hard}} = C_{t_i},
\end{equation}
\begin{equation}\label{eq:ab_hard_mask_wsoft}
    W_{t_i}^{\mathrm{soft}} = C_{t_i}.
\end{equation}
This ablation examines whether the performance gain comes merely from a roughly correct editable region, or from modeling this region as a continuous and smooth spatial support field.

\paragraph{No-transition-region variant.}
The no-transition-region variant preserves the soft mask itself but removes the explicitly constructed transition ring. In the full method, the transition region is obtained as:
\begin{equation}\label{eq:ab_ring_full_r}
    R_{t_i} = \mathrm{clamp}(D_{t_i} - C_{t_i}, 0, 1),
\end{equation}
and the hard support map is:
\begin{equation}\label{eq:ab_ring_full_whard}
    W_{t_i}^{\mathrm{hard}} = \mathrm{clamp}\!\left(M_{t_i} + w_{\mathrm{ring}} R_{t_i} (1-M_{t_i}), 0, 1\right).
\end{equation}
In this ablation, the transition region is removed:
\begin{equation}\label{eq:ab_no_ring_r}
    R_{t_i} = 0,
\end{equation}
and the support maps are defined as:
\begin{equation}\label{eq:ab_no_ring_whard}
    W_{t_i}^{\mathrm{hard}} = M_{t_i},
\end{equation}
\begin{equation}\label{eq:ab_no_ring_wsoft}
    W_{t_i}^{\mathrm{soft}} = s + (1-s) W_{t_i}^{\mathrm{hard}}.
\end{equation}
This variant isolates the contribution of the transition region as a boundary-smoothing mechanism.

\paragraph{No-temporal-accumulation variant.}
The no-temporal-accumulation variant disables the temporal accumulation mechanism. In the full method, the accumulated states are:
\begin{equation}\label{eq:ab_accum_full_soft}
    \bar W_{t_i}^{\mathrm{soft}} = \max\!\left(\bar W_{t_{i+1}}^{\mathrm{soft}}, W_{t_i}^{\mathrm{soft}}\right),
\end{equation}
\begin{equation}\label{eq:ab_accum_full_hard}
    \bar W_{t_i}^{\mathrm{hard}} = \max\!\left(\bar W_{t_{i+1}}^{\mathrm{hard}}, W_{t_i}^{\mathrm{hard}}\right),
\end{equation}
\begin{equation}\label{eq:ab_accum_full_c}
    \bar C_{t_i} = \max\!\left(\bar C_{t_{i+1}}, C_{t_i}\right),
\end{equation}
\begin{equation}\label{eq:ab_accum_full_r}
    \bar R_{t_i} = \max\!\left(\bar R_{t_{i+1}}, R_{t_i}\right).
\end{equation}
When temporal accumulation is removed, the model only uses the instantaneous support state at each time step:
\begin{equation}\label{eq:ab_no_accum_soft}
    \bar W_{t_i}^{\mathrm{soft}} \leftarrow W_{t_i}^{\mathrm{soft}},
\end{equation}
\begin{equation}\label{eq:ab_no_accum_hard}
    \bar W_{t_i}^{\mathrm{hard}} \leftarrow W_{t_i}^{\mathrm{hard}},
\end{equation}
\begin{equation}\label{eq:ab_no_accum_c}
    \bar C_{t_i} \leftarrow C_{t_i},
\end{equation}
\begin{equation}\label{eq:ab_no_accum_r}
    \bar R_{t_i} \leftarrow R_{t_i}.
\end{equation}
This experiment verifies whether SAM-Flow benefits from treating the editable region as a stable accumulated spatio-temporal support region.

\paragraph{No-source-anchored-projection variant.}
The no-source-anchored-projection variant keeps the same mask projection formula but replaces the anchor from the source-image latent with the latent state from the previous time step. In the full method, both anchors are set to the source latent:
\begin{equation}\label{eq:ab_source_anchor_full_model}
    a_{t_i}^{\mathrm{model}} = z^{\mathrm{src}},
\end{equation}
\begin{equation}\label{eq:ab_source_anchor_full_vis}
    a_{t_i}^{\mathrm{vis}} = z^{\mathrm{src}}.
\end{equation}
In this ablation variant, the anchor is replaced with the previous latent state:
\begin{equation}\label{eq:ab_source_anchor_removed_model}
    a_{t_i}^{\mathrm{model}} = z_{t_i}^{\mathrm{model}},
\end{equation}
\begin{equation}\label{eq:ab_source_anchor_removed_vis}
    a_{t_i}^{\mathrm{vis}} = z_{t_i}^{\mathrm{vis}}.
\end{equation}
The model-state update formula is still preserved:
\begin{equation}\label{eq:ab_prev_anchor_model}
    z_{t_{i-1}}^{\mathrm{model}} = a_{t_i}^{\mathrm{model}} + \rho \,\bar W_{t_i}^{\mathrm{soft}} \odot \left(z_{t_{i-1}}^{\mathrm{edit}} - a_{t_i}^{\mathrm{model}}\right).
\end{equation}
For the visible state, the update is:
\begin{equation}\label{eq:ab_prev_anchor_vis}
    z_{t_{i-1}}^{\mathrm{vis}} = \bar C_{t_i} \odot z_{t_{i-1}}^{\mathrm{core}} + \bar R_{t_i}(1-\bar C_{t_i}) \odot z_{t_{i-1}}^{\mathrm{ring}} + \mathrm{outer}_{t_i} \odot a_{t_i}^{\mathrm{vis}}.
\end{equation}
Compared with the full method, the reference is no longer the fixed source image. The background region is therefore no longer explicitly pulled back to the source-image latent trajectory, but only constrained around a previous state that may already have been modified.

\begin{figure*}[ht]
    \centering
    \includegraphics[width=\linewidth]{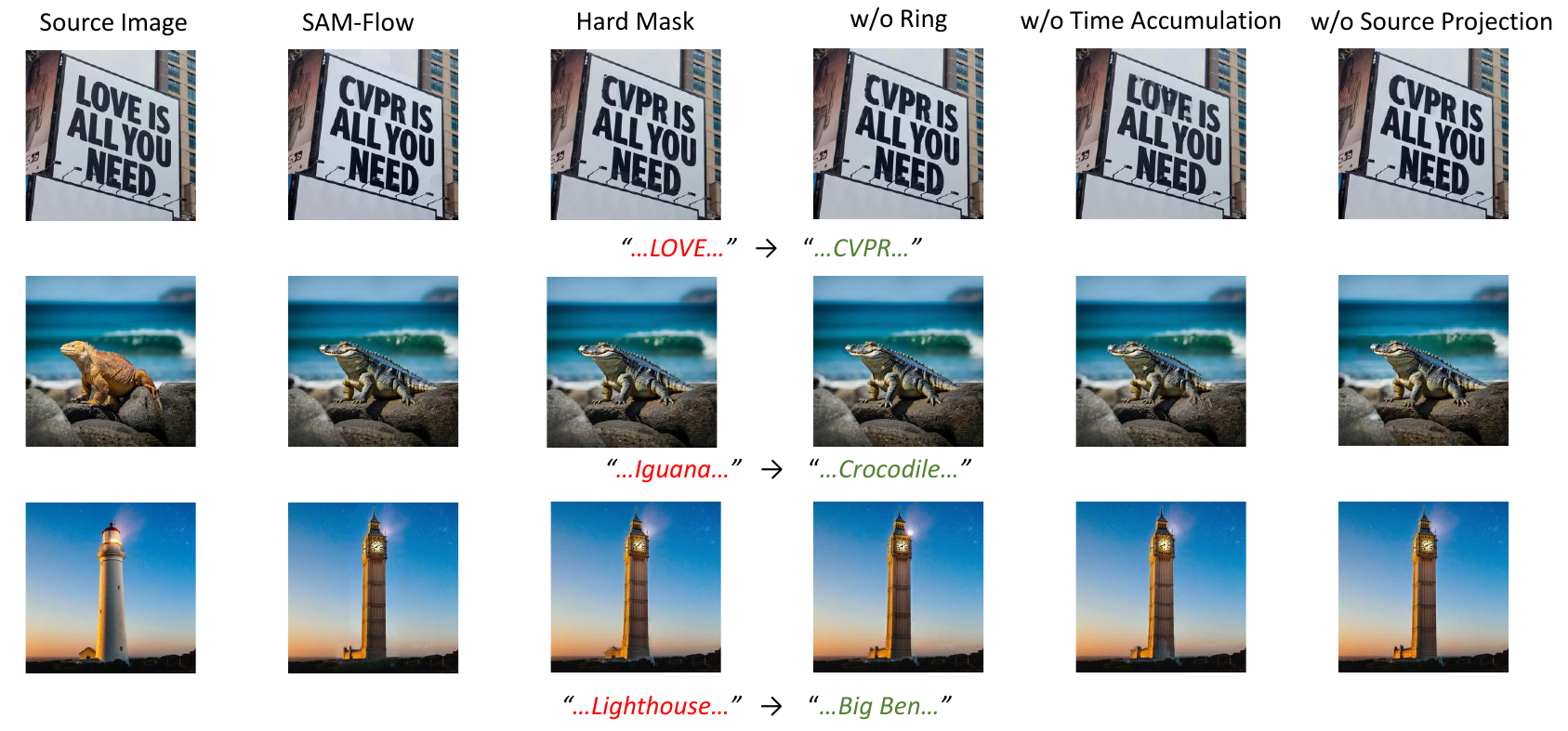}
    \caption{Qualitative results of the ablation study. The red text indicates the source prompt, and the green text indicates the target prompt.}
    \label{fig4.7}
\end{figure*}

\begin{table*}[ht]
\centering
\small
\caption{Quantitative results of the ablation study.}
\label{tab4.3-5}
\begin{tabular}{llccccc}
\toprule
Backbone & Variant & CLIP-T $\uparrow$ & DINO $\uparrow$ & PSNR $\uparrow$ & SSIM $\uparrow$ & LPIPS $\downarrow$ \\
\midrule
\multirow{5}{*}{SD3}
& SAM-Flow & 0.344 & \textbf{0.706} & 29.31 & \textbf{0.925} & \textbf{0.031} \\
& Hard mask & \textbf{0.346} & 0.661 & 29.54 & 0.916 & 0.123 \\
& No transition region & 0.343 & 0.681 & \textbf{30.22} & 0.924 & 0.087 \\
& No temporal accumulation & 0.294 & 0.538 & 26.59 & 0.877 & 0.226 \\
& No source-anchored projection & 0.341 & 0.702 & 28.77 & 0.904 & 0.036 \\
\midrule
\multirow{5}{*}{FLUX}
& SAM-Flow & 0.340 & \textbf{0.689} & 28.02 & 0.902 & \textbf{0.028} \\
& Hard mask & \textbf{0.344} & 0.669 & 28.00 & 0.894 & 0.121 \\
& No transition region & 0.338 & 0.682 & \textbf{28.13} & \textbf{0.913} & 0.085 \\
& No temporal accumulation & 0.296 & 0.545 & 25.06 & 0.861 & 0.227 \\
& No source-anchored projection & 0.341 & 0.688 & 27.87 & 0.892 & 0.032 \\
\bottomrule
\end{tabular}
\end{table*}

The qualitative comparison of the four ablation variants is shown in Fig.~\ref{fig4.7}. The hard-mask variant and the no-transition-region variant weaken visual quality and boundary naturalness. The no-temporal-accumulation variant may lead to inaccurate masks and editing failure. The no-source-anchored-projection variant is more likely to produce background drift. The quantitative results in Table~\ref{tab4.3-5} further validate this trend: the full method achieves the best overall trade-off between semantic compliance and background preservation on both Stable Diffusion 3 and FLUX.

\end{document}